\documentclass[final]{cvpr}

\usepackage{times}
\usepackage{epsfig}
\usepackage{graphicx}
\usepackage{amsmath}
\usepackage{amssymb}
\usepackage{multirow}
\usepackage{tabularx}
\usepackage{mathrsfs}
\usepackage{marvosym}
\usepackage[pagebackref=true,breaklinks=true,colorlinks,bookmarks=false]{hyperref}

\def\cvprPaperID{65} % *** Enter the CVPR Paper ID here
\def\confYear{CVPR 2021 Workshop}
\begin{document}

%%%%%%%%% TITLE
\title{PP-MSVSR: Multi-Stage Video Super-Resolution}

\author{Lielin Jiang \thanks{Both authors contributed equally to this works.}, 
Na Wang \footnotemark[1], 
Qingqing Dang, 
% \textsuperscript{\Letter}, 
Rui Liu,
Baohua Lai\\
Baidu Inc.\\
{\tt\small \{jianglielin, wangna11, dangqingqing, liurui25, laibaohua\}@baidu.com}
% \and
}

\maketitle

%%%%%%%%% ABSTRACT
\begin{abstract}
Different from the Single Image Super-Resolution(SISR) task, the key for Video Super-Resolution(VSR) task is to make full use of complementary information across frames to reconstruct the high-resolution sequence.  
Since images from different frames with diverse motion and scene, accurately aligning multiple frames and effectively fusing different frames has always been the key research work of VSR tasks. 
To utilize rich complementary information of neighboring frames, in this paper, we propose a multi-stage VSR deep architecture, dubbed as PP-MSVSR,  with local fusion module, auxiliary loss and re-align module to refine the enhanced result progressively.
Specifically, in order to strengthen the fusion of features across frames in feature propagation, a local fusion module is designed in stage-1 to perform local feature fusion before feature propagation.
Moreover, we introduce an auxiliary loss in stage-2 to make the features obtained by the propagation module reserve more correlated information connected to the HR space, and introduce a re-align module in stage-3 to make full use of the feature information of the previous stage.
Extensive experiments substantiate that PP-MSVSR achieves a promising performance of Vid4 datasets, which achieves a PSNR of 28.13dB with only 1.45M parameters. And the PP-MSVSR-L exceeds all state of the art method on REDS4 datasets with considerable parameters. Code and models will be released in PaddleGAN\footnote{https://github.com/PaddlePaddle/PaddleGAN.}.

\end{abstract}

%%%%%%%%% BODY TEXT
\section{Introduction}

The task of video super-resolution (VSR) is to recover the corresponding high-resolution (HR) counterpart from a given low-resolution (LR) video. VSR technology has received greater interest from researchers in recent years as a result of the explosive growth of video data over the internet, and it has become one of the research spotlights. VSR is also an ill-posed problem same as single-image super-resolution (SISR). However, unlike 
SISR, VSR requires not only the attention of the corresponding low-resolution frame, but also the utilization of information from consecutive frames in video sequences.

With the development of CNN network, various methods have been proposed to address the problem. Some early methods continued the idea of SISR and did not make good use of continuous video frame information, which also led to unsatisfactory final results. In order to make better use of continuous video frame information, recent methods often include multiple modules, such as inter-frame fusion, alignment, propagation and reconstruction modules. 
According to whether the video frames are explicitly aligned, VSR method can be categorized into two main groups: methods  without alignment and methods with alignment.  
Although the unaligned method \cite{yi2019progressive, yi2021omniscient, isobe2020revisiting} has a simple network structure, the restoration result for large motion videos is usually poor. To tackle this problem, a few studies \cite{li2019fast, kim2019video} have used 3D convolution to capture the relation between frames. To exploit similar patches across frames, MuCAN \cite{10.1007/978-3-030-58607-2_20} employs a temporal multi-correspondence aggregation approach. These networks, on the other hand, usually have a huge number of parameters.
\begin{figure}[t]
  \centering
%   \fbox{\rule{0pt}{2in} \rule{0.9\linewidth}{0pt}}
   \includegraphics[width=0.90\linewidth]{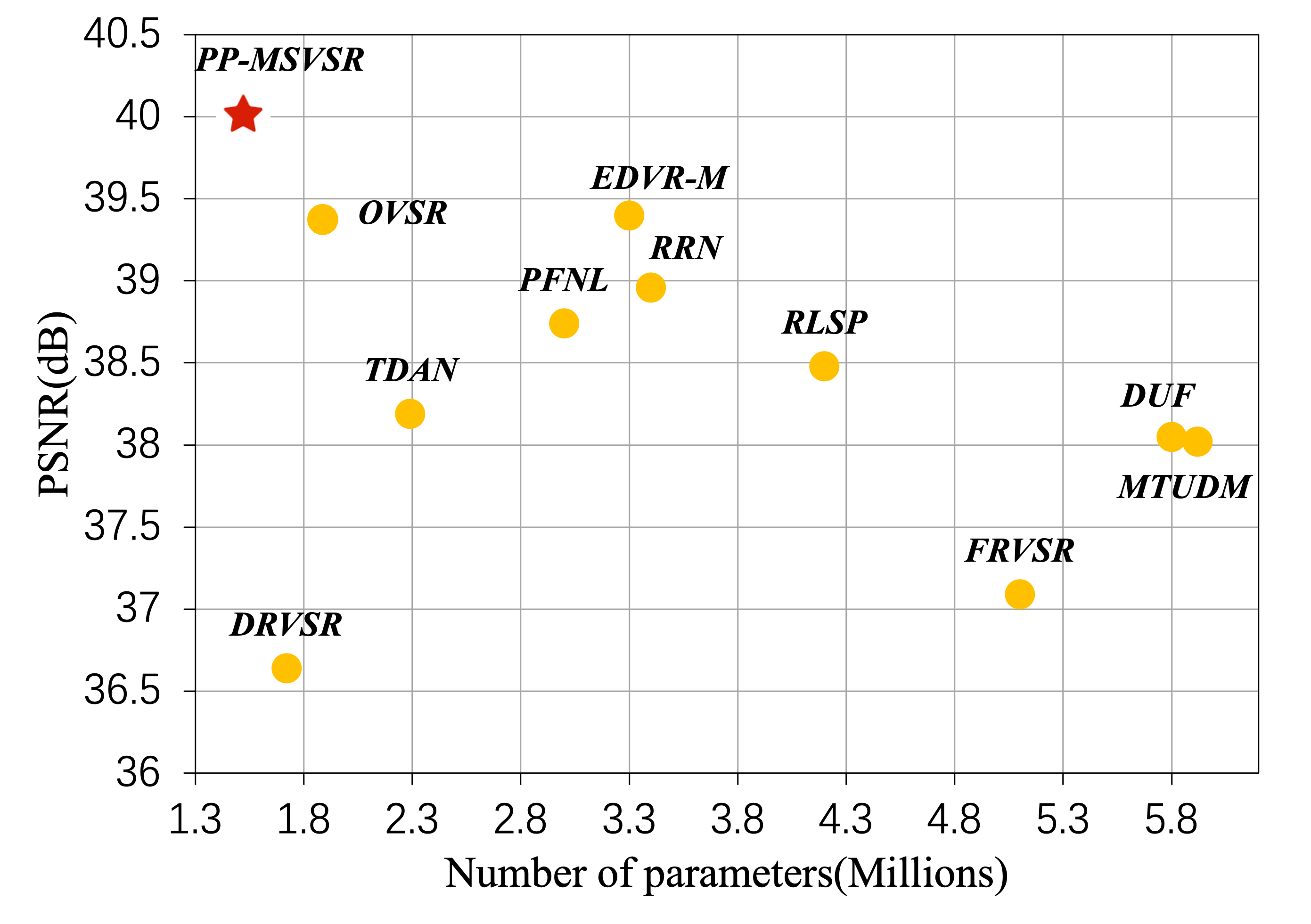}

   \caption{Video super-resolution on the UDM10 dataset with parameter less than 6M. Under different parameter capacities (x-axis), our multi-stage method performs better than the other method (PSNR on y-axis).}
   \label{fig:onecol}
\end{figure}

The sliding-window framework \cite{caballero2017real, yi2019multi, wang2019edvr, yi2019progressive, tian2020temporally} is a prevalent solution in which each frame in the video is restored using frames from a short temporal window. Despite the fact that the input consists of numerous consecutive video frames, sliding-window is comparable to treating the restoration of each video frame as a separate task for the full video sequence. A recurrent framework \cite{sajjadi2018frame, isobe2020video, isobe2020revisiting, chan2021basicvsr}, in contrast to the sliding-window framework, aims to exploit long-term interdependence by propagating latent features. In comparison to the sliding-window structure, these methods allow for a more compact model. As a result, the recurrent network's effect is highly dependent on the length of the video sequence. Recently, a few studies \cite{fuoli2019efficient, yan2019frame} have attempted to integrate the ideas of two frameworks. They either have a large number of parameters or the restoration result isn't very promising.

To address the aforementioned issues, we combine the idea of sliding-window framework and recurrent framework. A multi-stage framework was proposed to decompose a VSR pipeline into multiple sub-stage.

The main contributions of this paper are as follows:
\begin{quote}
(1) We propose a multi-stage network that combines the idea of sliding-window VSR and recurrent VSR , which makes full use of the information from the previous stage to gradually repair the video frame. 

(2) We design a local fusion module as stage-1, which performs local feature fusion before feature propagation to enhance the fusion of features across frames in feature propagation.

(3) We add an auxiliary loss in stage-2 to make the features obtained by the propagation module reserve more correlated information connected to the HR space. And at the latter stage, the propagation module will be aligned to this feature.

(4) We   propose   a   re-alignment   module(RAM) in stage-3  that  fusion  the  alignment  information  of the previous stage to further align the feature information of adjacent video frames.

\end{quote}
\section{Related Work}

% \subsection{Quality Enhancement}
\noindent
\textbf{Sliding-window Networks.} Some early VSR methods \cite{kappeler2016video, liao2015video} directly use the SISR architectures to reconstruct the video frame by frame.
As a result, the temporal infomation between adjacent frames isn't fully utilized.
To address this problem, recent methods designed a lot of sophisticated modules. PFNL \cite{yi2019progressive} proposes a progressive fusion block to fuse a short temporal frames directly. DUF \cite{jo2018deep} utilizes dynamic upsampling filters to
estimate motions implicitly. TDAN \cite{tian2020temporally} proposes a temporal deformable alignment network to adaptively align the neighborhood frame at the feature level. Based on TDAN, EDVR \cite{wang2019edvr}, the champion method of NTIRE2019 VSR competition, further utilizes deformable convolution (DCN) to design a pyramid, cascading and deformable (PCD) alignment module.

\noindent
\textbf{Recurrent networks.} Recurrent framework, in contrast to sliding-window framework, propagates the underlying information throughout the video sequence, demonstrating more powerful potential. FRVSR \cite{sajjadi2018frame} proposes a frame-recurrent network to utilize the previously inferred information. RSDN \cite{isobe2020video} uses structure-detail decomposition module and  hiddent-state adation module to propagate details of previous frames.  RRN \cite{isobe2020revisiting} uses a residual learning to stable training and boost the VSR performance. BasicVSR \cite{chan2021basicvsr} rethinks some components of previous methods and proposes a framework, which combines the bidirectional propagation and  a simple feature alignment module, obtains excellent results on the VSR task.

\noindent
\textbf{Multi-Stage Approaches.} Some image restoration methods \cite{fu2019lightweight,li2018recurrent,nah2017deep,ren2019progressive,zamir2021multi,zheng2019residual} in low level vision tasks utilize a subnetwork at each stage to restore the image in a gradual manner. Such a design is effective since it decomposes the challenging image restoration task into smaller easier subtasks. However, Unlike SISR, VSR often contains more complex components. How to integrate the features obtained by different components in the network at various stages is critical to the final recovery result. In EDVR \cite{wang2019edvr}, they train the two networks separately, and then use the two networks as separate stages to add their respective results and finally get the final result. Although the effect is improved in the end, this non-end-to-end approach is cumbersome and not practical enough. OVSR \cite{yi2021omniscient} combines the idea of sliding-widow framework and recurrent framework by using progressive fusion block from PFNL \cite{yi2019progressive}. Although it works well on specific datasets, the result may be decrease when dealing with large motion videos due to the lack of alignment modules.

\noindent
\textbf{Alignment.} More and more methods \cite{tian2020temporally, wang2019edvr, chan2021basicvsr, chan2021basicvsr++}, whether sliding-window frameworks or recurrent frameworks, have proved that the alignment of video frames is critical for VSR restore tasks. For approaches with alignment, there are two main methods: motion estimation and motion compensation (MEMC) or deformable convolution (DCN). For motion estimation, FRVSR \cite{sajjadi2018frame} and BasicVSR \cite{chan2021basicvsr} perform alignment at the image level using Spynet \cite{ranjan2017optical}.
For motion compensation, TDAN and EDVR use DCN to perform alignment at the feature level. Then Chan et al.\cite{chan2021basicvsr++} investigates the the principle behind DCN alignment and flow-based alignment, and proposes the offset-fidelity loss. On this basis, BasicVSR++ further proposes a flow-guided deformable alignment module, which combines the idea of two techniques, demonstrated great power.

\begin{figure*}[!t]
    \centering
    \includegraphics[width=1.0\linewidth]{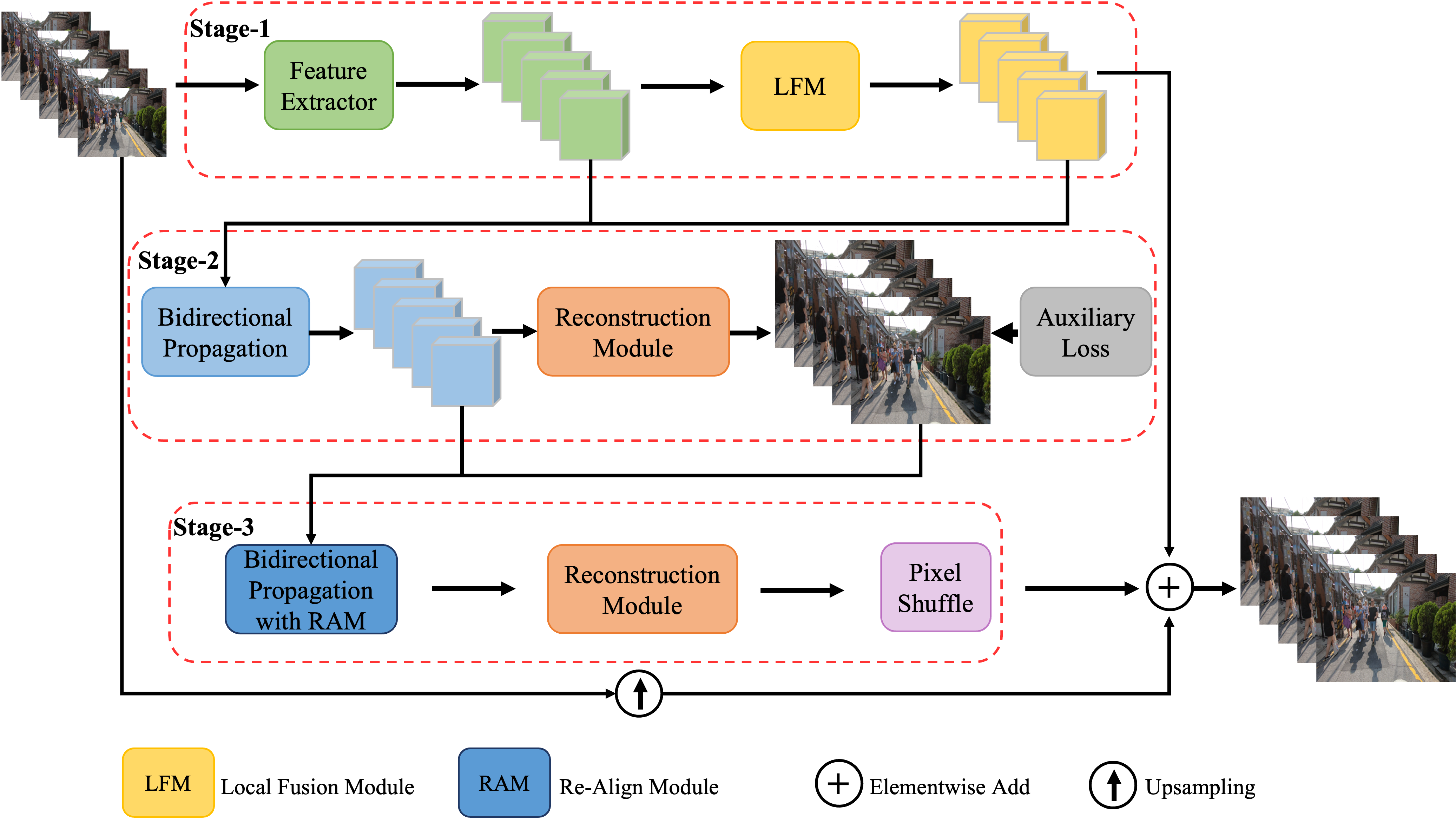}
    \caption{Workflow of PP-MSVSR, a multi-stage VSR deep architecture with local fusion module(LFM), auxiliary loss and re-align module(RAM).}
    \label{fig:overview}
\end{figure*}

\noindent
\textbf{Fusion.} Effective fusion of the extracted features is another critical step in the video restoration task. Some early methods use concatenate and convolutions to fuse all features from the previous module, which result in feature of adjacent video frames is not fully exploited. EDVR proposes a temporal and spatial attention (TSA) fusion module to integrate features from all aligned frames for subsequent restoration. BasicVSR uses bidirectional propagation to gradually fuse aligned feature. BasicVSR++ uses more aggressive bidirectional propagation arranged in a grid-like manner to repeatedly refine the features, improving expressiveness.

\section{Proposed Method}

\subsection{Multi-stage Network}
We first introduce the core idea of multi-stage video super-resolution (PP-MSVSR). As illustrated in Figure~\ref{fig:overview}, the network combines the idea of sliding-window VSR and recurrent VSR, and conducts the restoring task using a multi-stage strategy.

To be specific, let $ x_{i} $ be the input image, $ g_{i} $ be the feature extracted from $ x_{i} $ by multiple residual blocks, $ f_{LFM}^{i} $ be the feature after LFM described in Sec~\ref{subse:LFM},  
where $ i\in \left \{ 1,2,\cdots , N \right \} $ and $ N $ is the number of input frames.
The network extracts features from each video frame and performs a fusion of adjacent video frames by Local Fusion Module(FLM) in the Stage-1:
\begin{equation}
f_{LFM}^{i}  = \mathcal{FLM}(g_{i-1},g_{i},g_{i+1}).
\end{equation}

Inspired by the power of recurrent VSR network, we use the same structure as BasicVSR++ to fuse the information from different video frames and local merged features and then propagates the underlying information between each video frame at the second stage. In addition, we add a auxiliary loss to make feature more closed to HR space.

At the third stage, let $ f_{2}^{i} $ be the feature after Stage-2, let $ m_{2}^{i+1\to i} $ be the mask in Stage-2, let $ o_{2}^{i+1\to i} $ be the offset in Stage-2, $ f_{aligned}^{i} $ be the aligned feature in Stage-3.
We proposed a Re-Align module(RAM), which integrate offsets and masks of the stage-2 to facilitate precise motion compensation
\begin{equation}
f_{aligned}^{i+1} = \mathcal{RAM}(f_{2}^{i}, f_{2}^{i+1}, o_{2}^{i+1\to i}, m_{2}^{i+1\to i}).
\end{equation}
 The aligned features are sequentially fused, reconstructed, and upsampled, then the restored image is obtained.

\begin{figure}[t]
    \centering
    \includegraphics[width=1\linewidth]{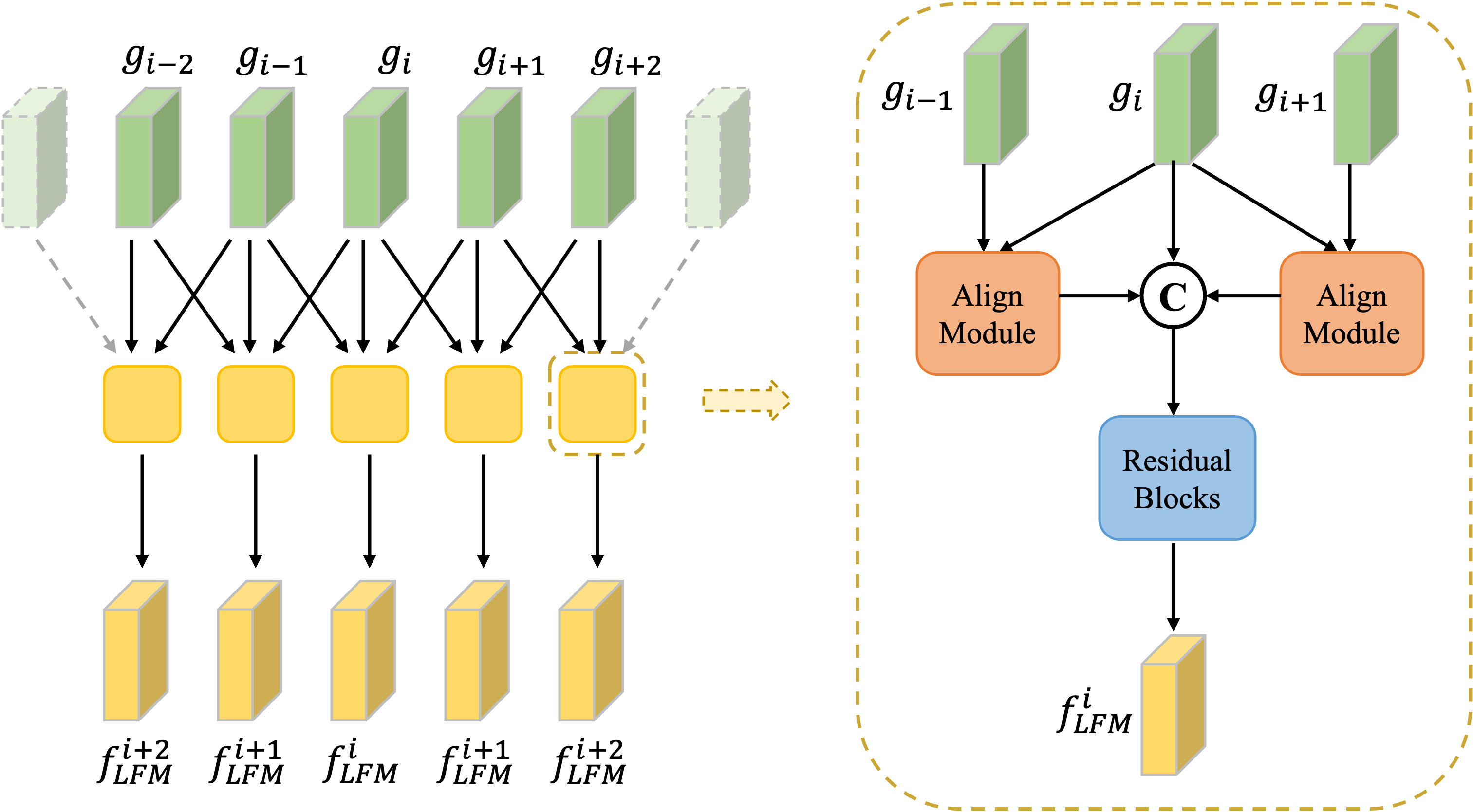}
    %\vspace{-1mm}
    \caption{LFM fusion module with local feature fusion .}
    \label{fig:LFM}
\end{figure}

% \begin{figure*}[!t]
% \end{figure*}

\subsection{Local Fusion Module}
\label{subse:LFM}

Inspired by the idea of sliding-window VSR, we designed a local fusion module in stage-1, denoted as LFM, to perform local feature fusion before feature propagation, which can strengthen cross-frame feature fusion in feature propagation. Specifically, as shown in Figure~\ref{fig:LFM}, the purpose of LFM is to let the features of the current frame fuse the information of its neighboring frames first, and then send the fused features to the propagation module.

 The LFM first aligns adjacent features through flow-guided deformable alignment introduced by BasicVSR++ \cite{chan2021basicvsr++}, and then merges the aligned adjacent features through concate and multiple residual blocks. When fusing 3 frames, the feature $ f_{LFM}^{i} $ can be expressed as
\begin{equation}
\label{form:LFM}
f_{LFM}^{i}  = \mathcal{C}(\mathcal{A}_{FGD} (g_{i-1},g_{i}),g_{i},\mathcal{A}_{FGD}(g_{i+1}, g_{i})).
\end{equation}
$ \mathcal{A}_{FGD} $ is the flow-guided deformable alignment
\begin{equation}
\hat{g}_{i-1}  = \mathcal{A}_{FGD} (g_{i-1},g_{i}) = \mathcal{D}(g_{i-1}, o_{i\to i-1}, m_{i\to i-1}),
\end{equation}
where $ \mathcal{D} $ denotes deformable convolution. $ o_{i\to i-1} $ denotes offset between $ g_{i-1} $ and $ g_{i} $, which combines the optical flow calculated by the SPynet \cite{ranjan2017optical} network and the offset calculated by deformable convolution. And $ m_{i\to i-1} $ denotes modulation masks.
$ \mathcal{C} $ is concate and a stack of residual blocks
\begin{equation}
\mathcal{C}(\hat{g}_{i-1},g_{i},\hat{g}_{i+1}) = R(concat(\hat{g}_{i-1},g_{i},\hat{g}_{i+1})).
\end{equation}
The fused features $ f_{LFM}^{i} $ will be sent to the propagation module in the stage-2.

%When fusing 5 frames, the second-order flow-guided deformable alignment introduced by BasicVSR++ cite{.....} is used to do align
%\begin{equation}
%f_{LFM}^{i}  = \mathcal{C}(\mathcal{A}_{FGD} (g_{i-2},g_{i-1},g_{i}),g_{i},\mathcal{A}_{FGD}(g_{i+2},g_{i+1}, g_{i})).
%\end{equation}
%where $ \mathcal{A}_{FGD} $ is the second-order flow-guided deformable alignment
%\begin{equation}
%\mathcal{A}_{FGD} (g_{i-2},g_{i-1},g_{i}) = \mathcal{D}(concat(g_{i-2},g_{i-1}), o_{i}, m_{i}),
%\end{equation}
%where $ o_{i} $ combines $ o_{i\to i-1} $ and $ o_{i\to i-2} $, $ m_{i} $ combines $ m_{i\to i-1} $ and $ m_{i\to i-2} $.

% We use BasicVSR \cite{chan2020basicvsr} without the pixel-shuffle \cite{shi2016real} layer as our base model of EnhanceNet in both stage-1 and stage-2. 

\subsection{Auxiliary Loss}

In the stage-2, we use the bidirectional propagation structure to propagate and fuse the information from different video frame, and use the flow-guided deformable alignment to align the features.

In addition, we add an auxiliary loss to make features more closed to HR space. 
To be specific, let $ f_{stage2}^{i} $ be the feature after propagation in stage-2. Then add auxiliary loss after upsampling $ f_{stage2}^{i} $
\begin{equation}
AuxLoss = \frac{1}{N}  {\textstyle \sum_{i=0}^{N}}  \sqrt{\left \| Up(f_{stage2}^{i}) - Y_{gt}^{i}   \right \| ^{2} + \varepsilon } ,
\end{equation}
where $ Up $ denotes upsampling and $ Y_{gt}^{i} $ denotes the ground truth.

\begin{table*}\small
\centering

\begin{tabularx}{0.88\textwidth}{lccccccccccc}
\hline
 Method & Parameter (M)   & FLOPs (G) & REDS4\cite{tao2017detail} & UDM10\cite{yi2019progressive} & Vimeo-90K-T \cite{xue2019video} & Vid4 \cite{liu2013bayesian}\\
\hline
Bicubic &- &-& 26.14/0.7292& 28.47/0.8253 &31.30/0.8687 &21.80/0.5246 \\
DRVSR\cite{tao2017detail} & 1.72 & 415 & - & 36.64/0.9472 & - & 25.52/0.7600 \\
FRVSR\cite{sajjadi2018frame} & 5.1 & 348 & - & 37.09/0.9522 & 35.64/0.9319 & 26.69/0.8103 \\
MTUDM\cite{yi2019multi} & 5.92 & 1672 & - & 38.02/0.9589 & - & 26.57/0.7989 \\
DUF\cite{jo2018deep}& 5.8 & 2348 & 28.63/0.8251 & 38.05/0.9586 & - & 27.34/0.8327 \\
EDVR-M\cite{wang2019edvr} & 3.3 & 200 & 30.53/0.8699 & 39.40/0.9663 & 37.33/0.9484 & 27.45/0.8406 \\
PFNL\cite{yi2019progressive} & 3.0 & 940 & 29.63/0.8502 & 38.74/0.9627 & - & 27.40/0.8384 \\
RLSP\cite{fuoli2019efficient} & 4.2 & 320 & - & 38.48/0.9606 & 36.49/0.9403 & 27.48/0.8388 \\
TDAN\cite{tian2020temporally} & 2.29 & 558 & - & 38.19/0.9586 & - & 26.86/0.8140 \\
RRN\cite{isobe2020revisiting} & 3.4 & - & - & 38.96/0.9644 & - & 27.69/0.8488 \\
OVSR\cite{yi2021omniscient} & 1.89 & 110 & - & 39.37/0.9673 & - & 27.99/0.8599 \\
\hline
PP-MSVSR & 1.45 & 111 & \textbf{31.25/0.8884} & \textbf{40.06/0.9699} & \textbf{37.54/0.9499} & \textbf{28.13/0.8604} \\
\hline
% \end{tabular}
\end{tabularx}
\caption{ Quantitative comparison (PSNR/SSIM). All results are calculated on Y-channel except REDS4 \cite{tao2017detail} (RGB-channel). Bold indicate the best performance. The FLOPs is computed on an LR size of 180×320. A 4× upsampling is performed following previous studies. Blanked entries correspond to results not reported in previous works.}
\label{tab:quantitative-1}
\end{table*}

\begin{table}
\centering
\begin{tabularx}{0.43\textwidth}{ccc}
\hline
Method & Parameter(M)& REDS \\
\hline
EDVR\cite{wang2019edvr} & 20.6 & 31.09/0.8800 \\
Iconvsr\cite{chan2021basicvsr} & 8.7& 31.67/0.8948 \\
BasicVSR++\cite{chan2021basicvsr++}& 7.3 & 32.39/0.9069 \\
\hline
PP-MSVSR-L & 7.4& \textbf{32.53/0.9083} \\
\hline
\end{tabularx}
\caption{ Quantitative comparison (PSNR/SSIM). All results are calculated on RGB-channel. Bold indicate the best performance. A 4× upsampling is performed following previous studies. }
\label{tab:quantitative-2}
\end{table}

\begin{table}
\centering
\begin{tabularx}{0.44\textwidth}{ccccc}
% \begin{tabular}{p{2.9cm}p{0.6cm}p{0.6cm}p{0.6cm}p{1.9cm}}
\hline
Method & A & B & C & PP-MSVSR \\
\hline
RAM & &\checkmark &\checkmark & \checkmark\\
LFM & & &\checkmark & \checkmark\\
Aux-Loss & & & & \checkmark  \\
\hline
PSNR &31.01&31.07&31.19&31.25\\
\hline
\end{tabularx}
\caption{ Ablation experiments of the components. Each component
brings significant improvements in PSNR, verifying their effectiveness.}
\label{tab:ablation}
\end{table}

\begin{figure}[t]
  \centering
   \includegraphics[width=0.98\linewidth]{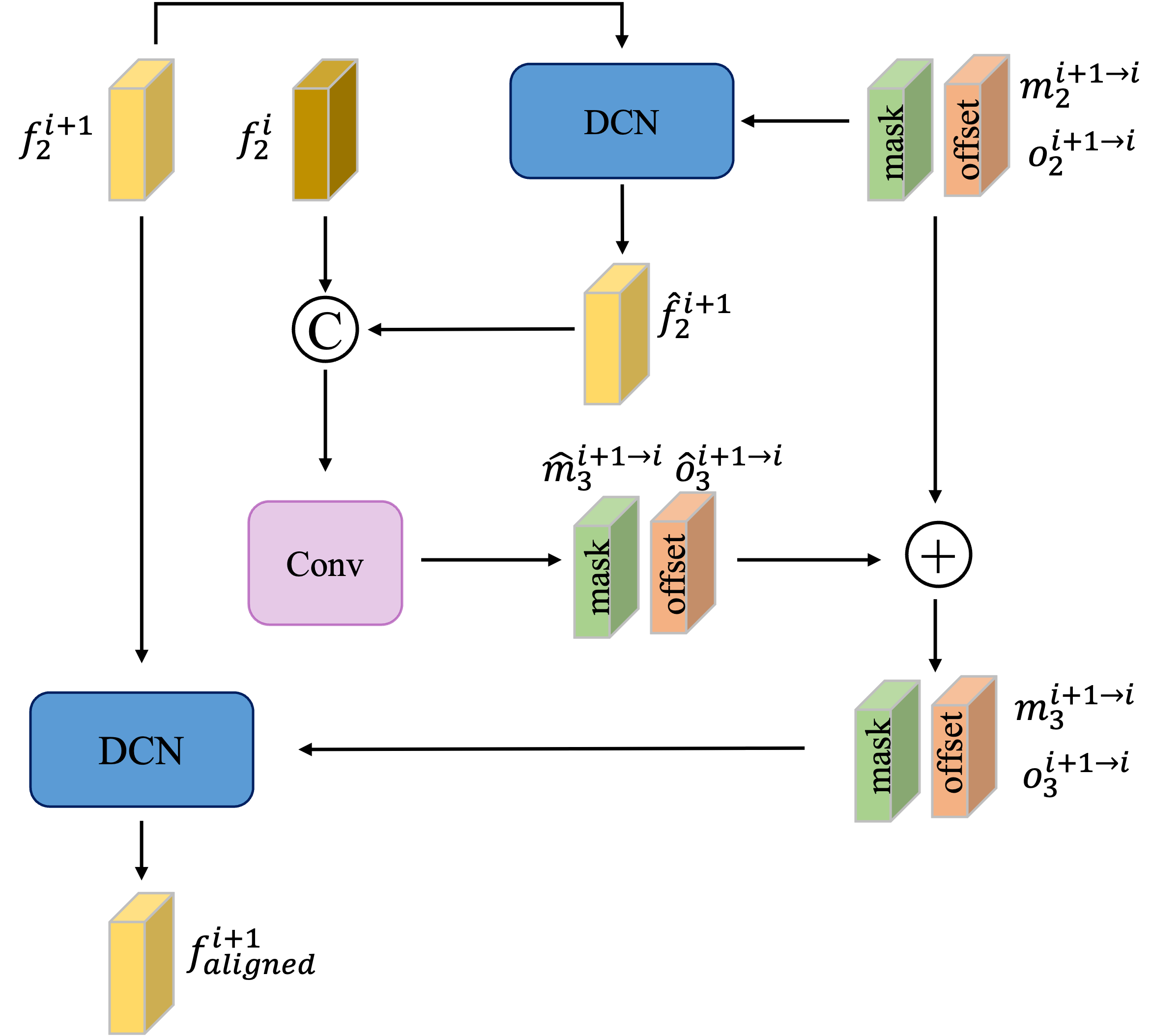}

   \caption{RAM align module with pre-align and align.}
   \label{fig:RAM}
\end{figure}

\subsection{Re-Align Module} %to be rename

Different from SISR,  in order to better integrate the information of adjacent frames, VSR usually aligns adjacent frames with the current frame. In some large motion video restoration tasks, the role of alignment is particularly obvious. In the process of using a bidirectional recurrent neural network, there are often multiple identical alignment operations. In order to make full use of the results of the previous alignment operations, we propose a Re-Align Module, denoted as RAM, that can utilize the previously aligned parameters and achieve a better alignment result. 

To be specific, as showed in Figure~\ref{fig:RAM}, previous mask and offset is used to pre-align the features
\begin{equation}
\hat{f} _{2}^{i+1} = \mathcal{D} (f_{2}^{i+1}, o_{2}^{i+1\to i}, m_{2}^{i+1\to i}),
\end{equation}
where $ \hat{f} _{2}^{i+1} $ denotes the pre-aligned features, $ \mathcal{D} $ denotes deformable convolution, $ m_{2}^{i+1\to i} $ and $ o_{2}^{i+1\to i} $ denotes the mask and offset in Stage-2. 
The aligned features are then concatenated to produce residual DCN offsets
\begin{equation}
\hat{o} _{3}^{i+1\to i}, \hat{m} _{3}^{i+1\to i} = Conv(concat (\hat{f} _{2}^{i+1}, f _{2}^{i}),
\end{equation}
Finally, two offsets are used for feature alignment
\begin{equation}
\hat{f} _{aligned}^{i+1} = \mathcal{D} (f_{2}^{i+1}, o_{2}^{i+1\to i}+\hat{o} _{3}^{i+1\to i}, m_{2}^{i+1\to i}+\hat{m} _{3}^{i+1\to i}).
\end{equation}
Then the aligned features are merged to reconstruct the restored image.

\subsection{PP-MSVSR-L}
Although the PP-MSVSR with the multi-stage architecture could achieve state-of-the-art performance in the models with the same parameter level, the performance is not the best among all the existing VSR models. In this paper, we also introduce a large VSR model guided by PP-MSVSR, dubbed as PP-MSVSR-L. 

Specifically, considering that BasicVSR++ is currently the most advanced method in VSR tasks, we increase the number of feature channels, input frames, backbone blocks and reconstruction blocks of the PP-MSVSR model to make the parameters of the new large VSR model consistent with BasicVSR++, and then our PP-MSVSR-L is obtained.
The experimental results show that the performance of PP-MSVSR-L exceeds that of BasicVSR++, and achieves state-of-the-art performance.

\section{Experiments}

\subsection{Datasets}

We use the two prevalent datasets for training and testing: REDS\cite{tao2017detail} and Vimeo-90K \cite{xue2019video}. REDS is a high-quality (720p) video dataset proposed in the NTIRE19 Competition. It contains 240 training clips, 30 validation clips and 30 testing clips (each with 100 consecutive frames).  To be consistent with previous methods, we use REDS4 as our test datasets, and the remaining clips are used for training. For Vimeo-90K, which is a widely used datasets in VSR task, we use Vid4 \cite{liu2013bayesian}, UDM10\cite{yi2019progressive} , and Vimeo90K-T as test sets. All models are tested with 4× downsampling using Blur Downsampling (BD).  

\subsection{Implementation Details}

We follow the most hyper-parameters of experiment in BasicVSR++. We used Adam optimizer\cite{kingma2014adam} by setting $ \beta 1 = 0.9 $, $ \beta 2 = 0.99 $. Cosine Annealing learning rate scheduler\cite{loshchilov2016sgdr} was adopted to decay the learning rate from $ 2 \times 10^{- 4} $ to $ 2 \times 10^{- 7} $. The initial learning rate of the main network and the flow network are set to $ 2\times 10^{- 4} $ and $ 2\times 10^{- 5} $ respectively. The total number of iterations is 300K, and the weights of the flow network are fixed during the first 2,500 iterations. The batch size is 16 and the patch size of input LR frames is $ 64 \times 64 $. We use Charbonnier loss\cite{charbonnier1994two} since it has better performance over the conventional L1 Loss and MSE loss. Considering the parameters and FLOPs, we use a pre-trained flow network as our flow network, which is modified from SPYNet \cite{ranjan2017optical} for efficient purposes. More details of modified SPYNet can be referred to in the PaddleGAN. In order to prove the effectiveness of our method, we also trained a large model with considerable parameters.

\subsection{Quantitative Results}
We compare PP-MSVSR and PP-MSVSR-L with several state-of-the-art VSR methods with fewer parameters and more parameters, respectively.

\begin{figure*}[!t]
    \centering
    \includegraphics[width=0.95\linewidth]{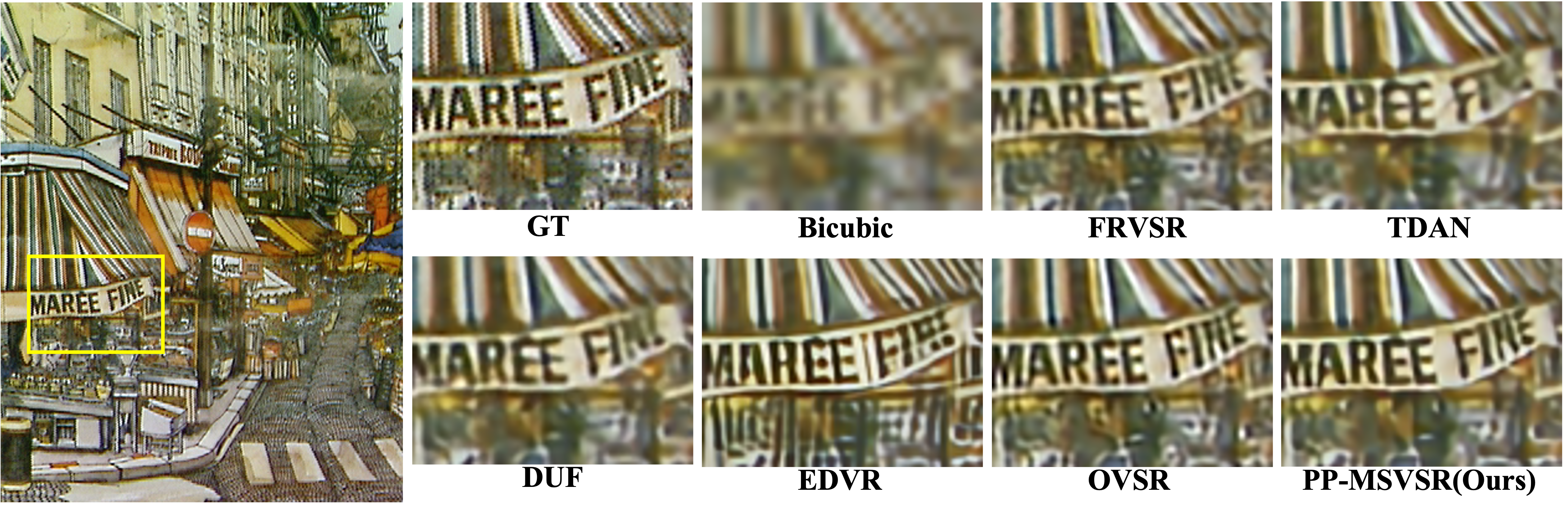}
    %\vspace{-1mm}
    \caption{Qualitative comparison on the \textbf{Vid4} dataset for $ 4\times  $ video SR}
    \label{fig:results-1}
\end{figure*}

%replace with reds
\begin{figure*}[!t]
    \centering
    \includegraphics[width=0.95\linewidth]{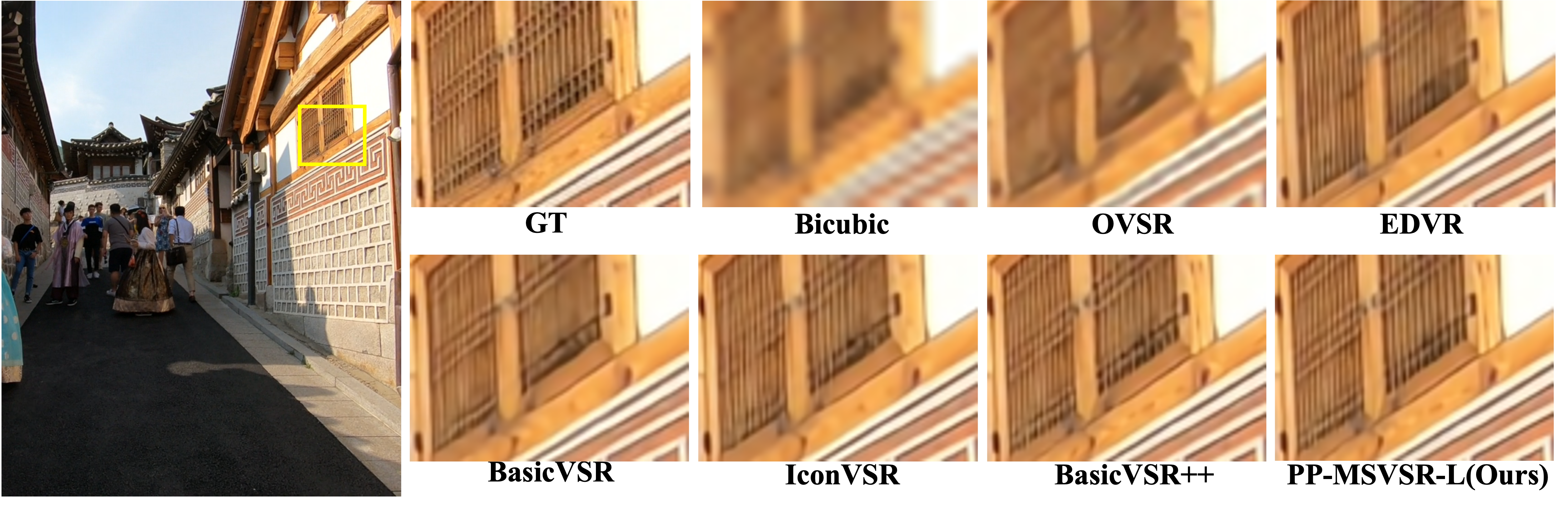}
    %\vspace{-1mm}
    \caption{Qualitative comparison on the \textbf{REDS4} dataset for $ 4\times  $ video SR}
    \label{fig:results-2}
\end{figure*}

\noindent
\textbf{PP-MSVSR.} We compare our PP-MSVSR method with eleven algorithms: DRVSR\cite{tao2017detail}, FRVSR\cite{sajjadi2018frame}, MTUDM\cite{yi2019multi}, DUF\cite{jo2018deep}, EDVR-M\cite{wang2019edvr}, PFNL\cite{yi2019progressive}, RLSP\cite{fuoli2019efficient}, TDAN\cite{tian2020temporally}, TGA\cite{tian2020temporally}, RRN\cite{isobe2020revisiting} and OVSR\cite{yi2021omniscient} on four testing datasets: REDS4\cite{tao2017detail}, UDM10\cite{yi2019progressive},  Vimeo-90K-T\cite{xue2019video} and Vid4\cite{liu2013bayesian}. In Table~\ref{tab:quantitative-1}, we provide parameter and FLOPs for each method, and provide PSNR and SSIM for each validation datasets. It can be seen that PP-MSVSR achieves state-of-the-art performance on all datasets with the lowest parameters. In particular, compared with EDVR-M, a sliding window method, our PP-MSVSR surpasses REDS-M on four datasets with nearly half parameters and FLOPs, and the PSNR on the REDS datasets is 0.7dB higher than EDVR-M, the PSNR on the Vid4 datasets is 0.7dB higher than EDVR-M.
And compared to light recurrent network RRN, the PSNR of PP-MSVSR with fewer parameters is 0.44dB higher on the Vid4 datasets, and 1.1dB higher on the UDM10 datasets.
Compared with the recent OVSR, our PP-MSVSR uses fewer parameters to achieve higher results.

\noindent
\textbf{PP-MSVSR-L.} Compared to state-of-the-art method EDVR\cite{wang2019edvr},  BasicVSR\cite{chan2021basicvsr} and BasicVSR++\cite{chan2021basicvsr++}, our PP-MSVSR-L achieved better results on both REDS datasets and Vid4 datasets, as listed in Table~\ref{tab:quantitative-2}. In particular, compared with the current best method BasicVSR++, PP-MSVSR-L achieved higher results with nearly the same parameters.

\subsection{Ablation Studies}

In order to understand the contribution of the proposed components, we start with the baseline and gradually insert the components. As can be seen from Table~\ref{tab:ablation}, each component has brought considerable improvements, RAM has increased by 0.06dB, LFM has increased by 0.12dB, Auxiliary Loss has increased by 0.06dB in RSNR.

%\subsection{Analysis and Discussions}

%Compared with quantitative results and qualitative results we can see that results with highest values of PSNR and SSIM are not the best human perceptual results. We also conduct user study to evaluate the temporal performance. For both validation and test videos, our results (fuse) and results (fuse + sharpen) have a better performance to reduce temporal flickering in smooth regions than results of EnhanceNet1 (lpips). Charbonnier loss \cite{lai2017deep} is helpful for generating smooth regions and lpips loss \cite{zhang2018unreasonable} is helpful for generating textual details. We can see that either model trained with only one loss function (Charbonnier loss or lpips loss) can not achieve a well enhanced video judged by human perception. Temporal flickering and lacking of textual details need to be consider simultaneously. Our proposed adaptive spatial-temporal fusion (ASTF) utilize both advantages of EnhanceNet1 (lpips) and EnhanceNet2 (cb + lpips) and have a better human perceptual video enhancement performance.

\section{Conclusion}
In this paper, we propose multi-stage VSR deep architecture PP-MSVSR based on BasicVSR++, and design LFM, auxiliary loss and RAM for its three stages respectively to refine the enhanced result  progressively. 
Experiments show the effectiveness of these three modules. In the small parameter model, our PP-MSVSR achieves state-of-the-art performance on all datasets with lowest parameters. And in the large model parameters, compared with the state-of-the-art method BasicVSR++, PP-MSVSR-L achieved higher results with nearly the same parameters. 

\section{Acknowledgments}
This work is supported by the National Key Research and Development Project of China (2020AAA0103503).

{\small

\bibliographystyle{ieee_fullname}

\begin{thebibliography}{10}\itemsep=-1pt

\bibitem{caballero2017real}
Jose Caballero, Christian Ledig, Andrew Aitken, Alejandro Acosta, Johannes
  Totz, Zehan Wang, and Wenzhe Shi.
\newblock Real-time video super-resolution with spatio-temporal networks and
  motion compensation.
\newblock In {\em Proceedings of the IEEE Conference on Computer Vision and
  Pattern Recognition}, pages 4778--4787, 2017.

\bibitem{chan2021basicvsr}
Kelvin~CK Chan, Xintao Wang, Ke Yu, Chao Dong, and Chen~Change Loy.
\newblock Basicvsr: The search for essential components in video
  super-resolution and beyond.
\newblock In {\em Proceedings of the IEEE/CVF Conference on Computer Vision and
  Pattern Recognition}, pages 4947--4956, 2021.

\bibitem{chan2021basicvsr++}
Kelvin~CK Chan, Shangchen Zhou, Xiangyu Xu, and Chen~Change Loy.
\newblock Basicvsr++: Improving video super-resolution with enhanced
  propagation and alignment.
\newblock {\em arXiv preprint arXiv:2104.13371}, 2021.

\bibitem{charbonnier1994two}
Pierre Charbonnier, Laure Blanc-Feraud, Gilles Aubert, and Michel Barlaud.
\newblock Two deterministic half-quadratic regularization algorithms for
  computed imaging.
\newblock In {\em Proceedings of 1st International Conference on Image
  Processing}, volume~2, pages 168--172. IEEE, 1994.

\bibitem{fu2019lightweight}
Xueyang Fu, Borong Liang, Yue Huang, Xinghao Ding, and John Paisley.
\newblock Lightweight pyramid networks for image deraining.
\newblock {\em IEEE transactions on neural networks and learning systems},
  31(6):1794--1807, 2019.

\bibitem{fuoli2019efficient}
Dario Fuoli, Shuhang Gu, and Radu Timofte.
\newblock Efficient video super-resolution through recurrent latent space
  propagation.
\newblock In {\em 2019 IEEE/CVF International Conference on Computer Vision
  Workshop (ICCVW)}, pages 3476--3485. IEEE, 2019.

\bibitem{isobe2020video}
Takashi Isobe, Xu Jia, Shuhang Gu, Songjiang Li, Shengjin Wang, and Qi Tian.
\newblock Video super-resolution with recurrent structure-detail network.
\newblock In {\em European Conference on Computer Vision}, pages 645--660.
  Springer, 2020.

\bibitem{isobe2020revisiting}
Takashi Isobe, Fang Zhu, Xu Jia, and Shengjin Wang.
\newblock Revisiting temporal modeling for video super-resolution.
\newblock {\em arXiv preprint arXiv:2008.05765}, 2020.

\bibitem{jo2018deep}
Younghyun Jo, Seoung~Wug Oh, Jaeyeon Kang, and Seon~Joo Kim.
\newblock Deep video super-resolution network using dynamic upsampling filters
  without explicit motion compensation.
\newblock In {\em Proceedings of the IEEE conference on computer vision and
  pattern recognition}, pages 3224--3232, 2018.

\bibitem{kappeler2016video}
Armin Kappeler, Seunghwan Yoo, Qiqin Dai, and Aggelos~K Katsaggelos.
\newblock Video super-resolution with convolutional neural networks.
\newblock {\em IEEE transactions on computational imaging}, 2(2):109--122,
  2016.

\bibitem{kim2019video}
Soo~Ye Kim, Jeongyeon Lim, Taeyoung Na, and Munchurl Kim.
\newblock Video super-resolution based on 3d-cnns with consideration of scene
  change.
\newblock In {\em 2019 IEEE International Conference on Image Processing
  (ICIP)}, pages 2831--2835. IEEE, 2019.

\bibitem{kingma2014adam}
Diederik~P Kingma and Jimmy Ba.
\newblock Adam: A method for stochastic optimization.
\newblock {\em arXiv preprint arXiv:1412.6980}, 2014.

\bibitem{li2019fast}
Sheng Li, Fengxiang He, Bo Du, Lefei Zhang, Yonghao Xu, and Dacheng Tao.
\newblock Fast spatio-temporal residual network for video super-resolution.
\newblock In {\em Proceedings of the IEEE/CVF Conference on Computer Vision and
  Pattern Recognition}, pages 10522--10531, 2019.

\bibitem{10.1007/978-3-030-58607-2_20}
Wenbo Li, Xin Tao, Taian Guo, Lu Qi, Jiangbo Lu, and Jiaya Jia.
\newblock Mucan: Multi-correspondence aggregation network for video
  super-resolution.
\newblock In Andrea Vedaldi, Horst Bischof, Thomas Brox, and Jan-Michael Frahm,
  editors, {\em Computer Vision -- ECCV 2020}, pages 335--351, Cham, 2020.
  Springer International Publishing.

\bibitem{li2018recurrent}
Xia Li, Jianlong Wu, Zhouchen Lin, Hong Liu, and Hongbin Zha.
\newblock Recurrent squeeze-and-excitation context aggregation net for single
  image deraining.
\newblock In {\em Proceedings of the European Conference on Computer Vision
  (ECCV)}, pages 254--269, 2018.

\bibitem{liao2015video}
Renjie Liao, Xin Tao, Ruiyu Li, Ziyang Ma, and Jiaya Jia.
\newblock Video super-resolution via deep draft-ensemble learning.
\newblock In {\em Proceedings of the IEEE International Conference on Computer
  Vision}, pages 531--539, 2015.

\bibitem{liu2013bayesian}
Ce Liu and Deqing Sun.
\newblock On bayesian adaptive video super resolution.
\newblock {\em IEEE transactions on pattern analysis and machine intelligence},
  36(2):346--360, 2013.

\bibitem{loshchilov2016sgdr}
Ilya Loshchilov and Frank Hutter.
\newblock Sgdr: Stochastic gradient descent with warm restarts.
\newblock {\em arXiv preprint arXiv:1608.03983}, 2016.

\bibitem{nah2017deep}
Seungjun Nah, Tae Hyun~Kim, and Kyoung Mu~Lee.
\newblock Deep multi-scale convolutional neural network for dynamic scene
  deblurring.
\newblock In {\em Proceedings of the IEEE conference on computer vision and
  pattern recognition}, pages 3883--3891, 2017.

\bibitem{ranjan2017optical}
Anurag Ranjan and Michael~J Black.
\newblock Optical flow estimation using a spatial pyramid network.
\newblock In {\em Proceedings of the IEEE conference on computer vision and
  pattern recognition}, pages 4161--4170, 2017.

\bibitem{ren2019progressive}
Dongwei Ren, Wangmeng Zuo, Qinghua Hu, Pengfei Zhu, and Deyu Meng.
\newblock Progressive image deraining networks: A better and simpler baseline.
\newblock In {\em Proceedings of the IEEE/CVF Conference on Computer Vision and
  Pattern Recognition}, pages 3937--3946, 2019.

\bibitem{sajjadi2018frame}
Mehdi~SM Sajjadi, Raviteja Vemulapalli, and Matthew Brown.
\newblock Frame-recurrent video super-resolution.
\newblock In {\em Proceedings of the IEEE Conference on Computer Vision and
  Pattern Recognition}, pages 6626--6634, 2018.

\bibitem{tao2017detail}
Xin Tao, Hongyun Gao, Renjie Liao, Jue Wang, and Jiaya Jia.
\newblock Detail-revealing deep video super-resolution.
\newblock In {\em Proceedings of the IEEE International Conference on Computer
  Vision}, pages 4472--4480, 2017.

\bibitem{tian2020temporally}
Yapeng Tian, Yulun Zhang, Yun Fu, and Chenliang~Xu TDAN.
\newblock temporally-deformable alignment network for video super-resolution.
  in 2020 ieee.
\newblock In {\em CVF Conference on Computer Vision and Pattern Recognition
  (CVPR)}, pages 3357--3366, 2020.

\bibitem{wang2019edvr}
Xintao Wang, Kelvin~CK Chan, Ke Yu, Chao Dong, and Chen Change~Loy.
\newblock Edvr: Video restoration with enhanced deformable convolutional
  networks.
\newblock In {\em Proceedings of the IEEE/CVF Conference on Computer Vision and
  Pattern Recognition Workshops}, pages 0--0, 2019.

\bibitem{xue2019video}
Tianfan Xue, Baian Chen, Jiajun Wu, Donglai Wei, and William~T Freeman.
\newblock Video enhancement with task-oriented flow.
\newblock {\em International Journal of Computer Vision}, 127(8):1106--1125,
  2019.

\bibitem{yan2019frame}
Bo Yan, Chuming Lin, and Weimin Tan.
\newblock Frame and feature-context video super-resolution.
\newblock In {\em Proceedings of the AAAI Conference on Artificial
  Intelligence}, pages 5597--5604, 2019.

\bibitem{yi2021omniscient}
Peng Yi, Zhongyuan Wang, Kui Jiang, Junjun Jiang, Tao Lu, Xin Tian, and Jiayi
  Ma.
\newblock Omniscient video super-resolution.
\newblock {\em arXiv preprint arXiv:2103.15683}, 2021.

\bibitem{yi2019progressive}
Peng Yi, Zhongyuan Wang, Kui Jiang, Junjun Jiang, and Jiayi Ma.
\newblock Progressive fusion video super-resolution network via exploiting
  non-local spatio-temporal correlations.
\newblock In {\em Proceedings of the IEEE/CVF International Conference on
  Computer Vision}, pages 3106--3115, 2019.

\bibitem{yi2019multi}
Peng Yi, Zhongyuan Wang, Kui Jiang, Zhenfeng Shao, and Jiayi Ma.
\newblock Multi-temporal ultra dense memory network for video super-resolution.
\newblock {\em IEEE Transactions on Circuits and Systems for Video Technology},
  30(8):2503--2516, 2019.

\bibitem{zamir2021multi}
Syed~Waqas Zamir, Aditya Arora, Salman Khan, Munawar Hayat, Fahad~Shahbaz Khan,
  Ming-Hsuan Yang, and Ling Shao.
\newblock Multi-stage progressive image restoration.
\newblock In {\em Proceedings of the IEEE/CVF Conference on Computer Vision and
  Pattern Recognition}, pages 14821--14831, 2021.

\bibitem{zheng2019residual}
Yupei Zheng, Xin Yu, Miaomiao Liu, and Shunli Zhang.
\newblock Residual multiscale based single image deraining.
\newblock In {\em bmvc}, page 147, 2019.

\end{thebibliography}
}

\end{document}